\documentclass[letterpaper, 10 pt, conference]{ieeeconf}  

\IEEEoverridecommandlockouts                              

\usepackage{siunitx}
\usepackage[dvipdfmx]{graphicx}
\usepackage{amsmath}
\usepackage{amsfonts}
\usepackage{amssymb}
\usepackage{url}
\usepackage{multirow}
\usepackage{subcaption}
\usepackage[switch]{lineno} 
\newcommand*\patchAmsMathEnvironmentForLineno[1]{
  \expandafter\let\csname old#1\expandafter\endcsname\csname #1\endcsname
  \expandafter\let\csname oldend#1\expandafter\endcsname\csname end#1\endcsname
  \renewenvironment{#1}
     {\linenomath\csname old#1\endcsname}
     {\csname oldend#1\endcsname\endlinenomath}}
\newcommand*\patchBothAmsMathEnvironmentsForLineno[1]{
  \patchAmsMathEnvironmentForLineno{#1}
  \patchAmsMathEnvironmentForLineno{#1*}}
\AtBeginDocument{
\patchBothAmsMathEnvironmentsForLineno{equation}
\patchBothAmsMathEnvironmentsForLineno{align}
\patchBothAmsMathEnvironmentsForLineno{flalign}
\patchBothAmsMathEnvironmentsForLineno{alignat}
\patchBothAmsMathEnvironmentsForLineno{gather}
\patchBothAmsMathEnvironmentsForLineno{multline}
}

\DeclareMathOperator*{\argmax}{arg\,max}

\title{\LARGE \bf
Improving Wind Resistance Performance of Cascaded PID Controlled Quadcopters using Residual Reinforcement Learning
}

\author{Yu Ishihara$^{1}$, Yuichi Hazama$^{2}$, Kousuke Suzuki$^{2}$, Jerry Jun Yokono$^{1}$, Kohtaro Sabe$^{2}$, and Kenta Kawamoto$^{1,3}$
\thanks{$^{1}$Sony Group Corporation, 1-7-1 Konan Minato-ku, Tokyo, 108-0075, Japan. Corresponding author: 
        {\tt\small yu.ishihara@sony.com}}%
\thanks{$^{2}$Aerosense Inc., 5-41-10 Koishikawa, Bunkyo-ku, Tokyo, 112-0002, Japan}%
\thanks{$^{3}$Sony Research Inc., 1-7-1 Konan Minato-ku, Tokyo, 108-0075, Japan}%
}

\begin{document}

\maketitle
\thispagestyle{empty}
\pagestyle{empty}

\begin{abstract}
	Wind resistance control is an essential feature for quadcopters to maintain their position to avoid
	deviation from target position and prevent collisions with obstacles.
	Conventionally, cascaded PID controller is used for the control of quadcopters
	for its simplicity and ease of tuning its parameters.
	However, it is weak against wind disturbances and the quadcopter can easily deviate from target position.
	In this work, we propose a residual reinforcement learning based approach to build a wind resistance controller of a quadcopter.
	By learning only the residual that compensates the disturbance,
	we can continue using the cascaded PID controller as the
	base controller of the quadcopter but improve its performance against wind disturbances.
	To avoid unexpected crashes and destructions of quadcopters,
	our method does not require real hardware for data collection and training.
	The controller is trained only on a simulator and directly applied to the
	target hardware without extra finetuning process.
	We demonstrate the effectiveness of our approach through various experiments
	including an experiment in an outdoor scene with wind speed greater than \SI{13}{m/s}.
	Despite its simplicity, our controller reduces the position deviation by approximately \SI{50}{\%}
	compared to the quadcopter controlled with the conventional cascaded PID controller.
	Furthermore, trained controller is robust and preserves its performance even though the quadcopter's
	mass and propeller's lift coefficient is changed between \SI{50}{\%} to \SI{150}{\%} from
	original training time.
\end{abstract}

\section{INTRODUCTION}

Quadcopters are coming to be used for a variety of applications such as delivery, filming, and surveying.
These applications often require a quadcopter to be used in an outdoor scene where
an unpredictable wind disturbances exist.
A quadcopter flying outdoors may collide with nearby obstacles or
deviate from target position when an unexpected gust of wind blows.
Therefore, there is a high demand on quadcopters with a controller that can stabilize and maintain
its position and orientation under wind disturbances.

Model-free cascaded PID controllers are widely used as a controller for quadcopters because of its simplicity
to implement and to tune their parameters~\cite{ICRA2015_Meier}. However, this controller is prone to wind disturbances
due to its controller design\cite{ICUAS2017_Suarez,IJRR2012_Mellinger}. Cascaded PID controller needs to wait for the convergence of
subsequent layers to reflect the higher layers control input (e.g. position control input).
Hence, the controller delays in react and deviates from target position when sudden external disturbances occur.
Existing research approaches this issue by replacing the cascaded PID controller.
However, these approaches require to model the system's dynamics~\cite{ICUAS2016_Bannwarth,ICUAS2018_Masse,ACC2022_Schmid,RAM2018_Sa,ICUAS2017_Suarez} or
learn a controller from data~\cite{ICRA2021_Sohege} for different quadcopters.
Therefore, the simplicity of cascaded PID controller is lost and makes it difficult to apply the controller on a variety of quadcopters.
Especially in industrial applications, it is often preferred to keep the simplicity of
PID controller but to make the system robust against wind disturbances.

In this work, we investigate for a method to build a robust controller against wind disturbances using
cascaded PID controller as a base controller of the system.
Specifically, we apply residual reinforcement learning~\cite{ICRA2019_Johannink} 
and make a policy model learn only the residual of control input that compensates the wind disturbances.
To avoid unexpected crashes and destructions of a quadcopter,
we train the controller solely on a simulator and directly apply the trained controller to the target hardware without
extra fine tuning in a real environment.
The training only requires approximately 12 hours wall-clock time.
We will demonstrate the effectiveness of our approach through various simulations and experiments including an experiment
in an outdoor scene with wind speed greater than \SI{13}{m/s}.
Despite its simplicity, the proposed method reduces the position deviation by approximately \SI{50}{\%} compared to the conventional cascaded PID controller.
It also preserves its performance even when the quadcopter's parameter (mass and propeller's lift-coefficient) changes
between \SI{50}{\%} to \SI{150}{\%} from original training time.

The contributions of this paper are as follows.
\begin{itemize}
	\item Proposal of wind resistance controller for quadcopters trained with residual reinforcement learning to
	      enable cascaded PID controller resistant to wind disturbances.
	\item Demonstration of the effectiveness of the proposed controller through simulations and experiments including
	      an experiment in an outdoor environment with wind speed greater than \SI{13}{m/s}.
\end{itemize}

\section{RELATED WORK}
\subsection{Wind Resistance Control of Quadcopters}
Wind resistant controller for quadcopters has been an active area of research for their high demands
on many commercial applications.
Previous approaches estimate the wind disturbance~\cite{ICUAS2016_Bannwarth, ACC2022_Schmid, RAM2018_Sa} or
apply robust control method~\cite{ICUAS2018_Masse,ICUAS2017_Suarez} to make the quadcopter's
controller resistant against wind disturbances. To enable such control, the quadcopter's dynamics
is estimated or determined in advance~\cite{ICUAS2016_Bannwarth,ICUAS2018_Masse,ACC2022_Schmid,RAM2018_Sa} and
control algorithms such as MPC~\cite{ICUAS2016_Bannwarth,ACC2022_Schmid,RAM2018_Sa},
$H_\infty$~\cite{ICUAS2018_Masse}, or $L_{1}$ adaptive control~\cite{ICUAS2017_Suarez} are applied to control the quadcopter under wind disturbances.
These approaches require identifying the quadcopter's system and customizing their control parameters
for each quadcopter. Hence, they are difficult to apply in case the system's dynamics changes drastically
during the operation (e.g. package delivery with a quadcopter).
Considering varieties of applications and quadcopters,
it is preferred to build a controller that can be used even though a quadcopter's system changes.
In this work, we incorporate residual reinforcement learning approach and build a wind resistant controller
for quadcopters. We will show that the learned controller can control the quadcopter
even when the quadcopter's parameter (mass and propeller's lift-coefficient) changes
between \SI{50}{\%} to \SI{150}{\%} from original training time.

\subsection{Reinforcement Learning for Robotic Applications}
Recently, reinforcement learning~\cite{RL2018_Sutton} has been used to train a wide variety of robotic applications
including quadcopters~\cite{ICRA2021_Taylor,ICRA2021_Li,SR2022_Miki,ICRA2018_Quillen, ICUAS2018_Vankadari}.
Since reinforcement learning enables to learn and optimize robot's behavior without knowing the underlying
dynamics of the system, it has been applied to applications that conventional model-based methods are
difficult to apply. Taylor et al. trained a bipedal robot to learn its locomotion from human motion capture data~\cite{ICRA2021_Taylor}.
Li et al. also trained a bipedal robot and realized a robust controller compared to conventional model-based controller that
performs a set of diverse and dynamic behaviors of bipedal robots~\cite{ICRA2021_Li}.
Miki et al. applied reinforcement learning to navigate a quadrupedal robot on different types of terrains
that their surface characteristics are difficult to model~\cite{SR2022_Miki}.
Reinforcement learning is also used for manipulators to learn grasping objects in a cluttered environment~\cite{ICRA2018_Quillen}.
Furthermore, Vankadari et al. trained a controller for autonomous landing of a quadcopter using reinforcement learning~\cite{ICUAS2018_Vankadari}.
In this paper, we will apply reinforcement learning to learn a wind resistant controller for quadcopters.
Since wind disturbances are difficult to model and predict in practice,
reinforcement learning was expected as a solution to realize such a controller.
We will demonstrate that the trained controller enables to cancel wind disturbances
and improve the quadcopter's performance compared to conventional cascaded PID-based controller.

\section{PRELIMINARIES}
We train our controller considering the problem as a standard reinforcement learning problem~\cite{RL2018_Sutton}.
Reinforcement learning is defined as a policy search in an environment modeled as a Markov decision process, defined by a tuple $(\mathcal{S},\mathcal{A},\rho_0,p,r)$.
Where $\mathcal{S}$, $\mathcal{A}$, $\rho_{0}$, $p$, $r$ denote the state space, action space,
initial state probability, transition probability, and reward function of the environment.
The objective of reinforcement learning is to find an optimal policy $\pi^{\ast}$ that maximizes the
following expected sum of discounted rewards
\begin{equation}
	\label{eq3-1}
	\pi^{\ast} = \argmax_{\pi} \mathbb{E}_{s_0\sim\rho_{0},s_{t>0}\sim p,a_t\sim\pi}[\sum_{t=0}^{\infty}\gamma^{t}r(s_t, a_t)].
\end{equation}
Here, $\gamma\in[0,1)$ is the discount factor, $s_t\in\mathcal{S}$ and $a_t\in\mathcal{A}$ represents the state and action at time $t$.
In this work, we used Soft Actor-Critic (SAC)~\cite{ICML2018_Haarnoja,Arxiv2018_Haarnoja} as the reinforcement learning algorithm
to train the policy $\pi$. In SAC, the policy is trained using a reward $r_{\text{target}}$ with entropy bonus defined as follows:
\begin{equation}
	\label{eq3-2}
	r(s_t, a_t) \triangleq r_{\text{target}}(s_t,a_t) + \alpha\mathcal{H}(\pi(\cdot|s_{t})).
\end{equation}
$\mathcal{H}(\pi(\cdot|s_t))$ denotes the entropy of the policy and $\alpha$ is a trade-off coefficient.
We used SAC because it is known to be robust and sample efficient for robotic tasks~\cite{Arxiv2018_Haarnoja}.
Furthermore, to take advantage of conventional cascaded PID controller, we incorporate the idea of residual reinforcement learning~\cite{ICRA2019_Johannink}.
Our policy $\pi$ computes action $a$ at each timestep as:
\begin{equation}
	\label{eq3-3}
	a = a_{\text{PID}} + a_{\text{RL}}.
\end{equation}
$a_{\text{PID}}$ is the output of cascaded PID controller and $a_{\text{RL}}$ is the action computed
using parameterized policy $\pi_{\theta}$ trained with SAC. $\theta$ is the parameter of the training policy.
Residual reinforcement learning enables the policy $\pi_{\theta}$ to focus on learning to cancel external disturbances
and avoids learning to fly the quadcopter from scratch.
Therefore, we expected that the training gets much easier and the training time reduces compared
to training a policy from scratch.

\section{METHOD}

\begin{figure}
	\centering
	\includegraphics[scale=0.07]{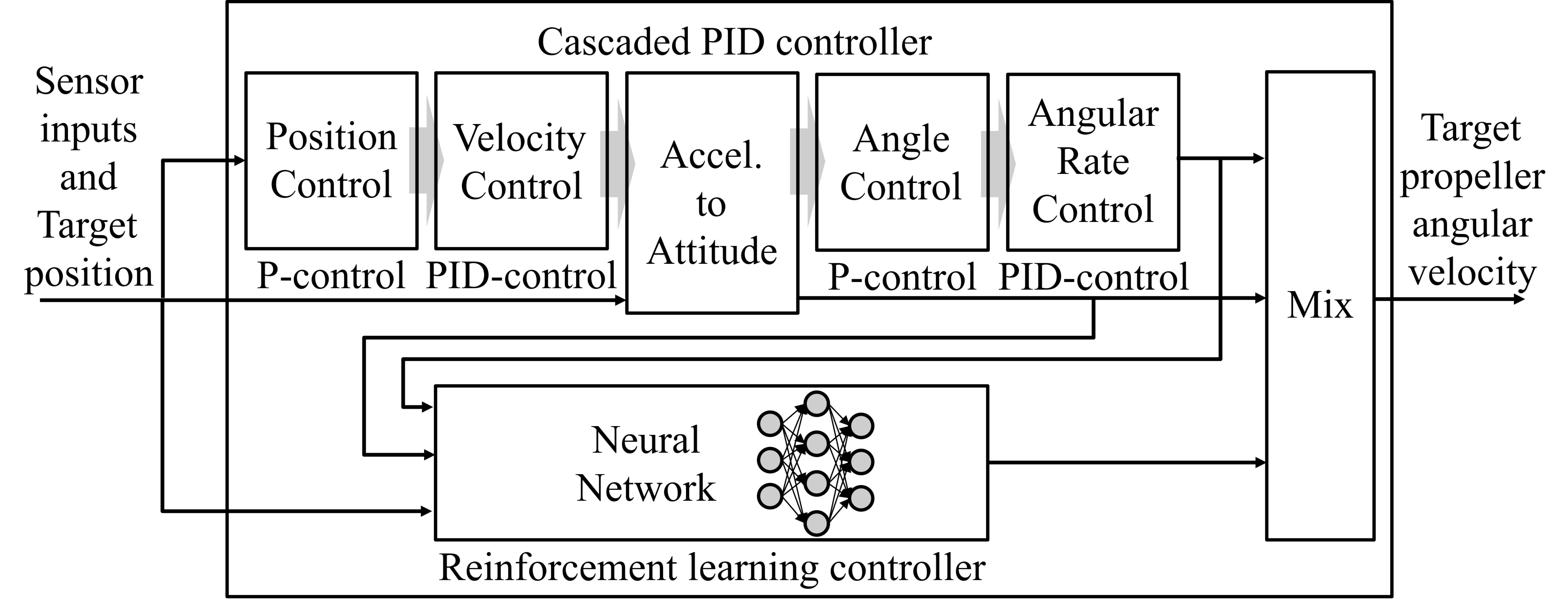}
	\caption{Architecture of proposed controller. PID controller's running frequency is between \SI{100}{Hz} to \SI{1000}{Hz}
		(Position control is around \SI{100}{Hz} but angular rate control is around \SI{1000}{Hz}).
		Reinforcement learning controller operates at \SI{10}{Hz}.}
	\label{fig4-1}
\end{figure}
Fig.~\ref{fig4-1} shows the overview of the proposed controller.
As mentioned in previous section, the quadcopter's controller consists of
a cascaded PID controller and a neural network-based controller which is trained using reinforcement learning algorithm.
Two controllers run in parallel and their output is mixed to compute the target angular velocity of each propeller.
In this work, we made use of cascaded PID controller also used in PX4~\cite{ICRA2015_Meier} as baseline controller. In the following subsections,
we will describe the design and the training method of the reinforcement learning controller in detail.

\subsection{Network Model and Controller Design}

We designed two neural network models $\pi_{\theta}$ and $Q_{\psi}$ to train the reinforcement learning controller.
$\theta$ and $\psi$ denotes the parameter of each neural network.
$\pi_{\theta}$ corresponds to the actor and $Q_{\psi}$ is the critic used in SAC algorithm.
$Q_{\psi}$ is only used during the training phase to update the parameter of controller $\pi_{\theta}$ with SAC.
\begin{table}[t]
	\caption{Network structure of $\pi_{\theta}$.}
	\label{tab4-1}
	\centering
	\begin{tabular}{|c|c|}
		\hline
		Input layer    & state input (68 dim)                       \\
		\hline
		Middle layer 1 & Fully-connected followed by relu (30 dim)  \\
		\hline
		Middle layer 2 & Fully-connected followed by relu (20 dim)  \\
		\hline
		Middle layer 3 & Fully-connected followed by relu (5 dim)   \\
		\hline
		Output layer   & Gaussian mean (4 dim) and variance (4 dim) \\
		\hline
	\end{tabular}
\end{table}
\begin{table}[t]
	\caption{Network structure of $Q_{\psi}$.}
	\label{tab4-2}
	\centering
	\begin{tabular}{|c|c|}
		\hline
		Input layer    & state and action inputs (68+4 dim)         \\
		\hline
		Middle layer 1 & Fully-connected followed by relu (400 dim) \\
		\hline
		Middle layer 2 & Fully-connected followed by relu (200 dim) \\
		\hline
		Middle layer 3 & Fully-connected followed by relu (100 dim) \\
		\hline
		Output layer   & Q-value (1 dim)                            \\
		\hline
	\end{tabular}\end{table}
Table~\ref{tab4-1} and Table~\ref{tab4-2} shows the structure of both networks.
$\pi_{\theta}$ is a small neural network which consists of four fully-connected layers.
We selected the number of intermediate neurons to enable running the network on quadcopter at \SI{10}{Hz}.
State input is a 68 dimensional vector which consists of following features:
\begin{itemize}
	\item history of relative position to target (3 dim$\times$3 steps)
	\item history of quadcopter's velocity (3 dim$\times$3 steps)
	\item history of relative angle to target (3 dim$\times$3 steps)
	\item history of quadcopter's angular velocity (3 dim$\times$3 steps)
	\item history of PID-controller's output (4 dim$\times$3 steps)
	\item history of RL-controller's output (4 dim$\times$5 steps)
\end{itemize}
Here, 1 step corresponds to \SI{0.1}{s}.
Final layer is split into two and used as the mean and the variance of a Gaussian distribution.
Action $\hat{a_{\text{RL}}}=(F^{\text{RL}}_{\text{thrust}},T^{\text{RL}}_{\text{roll}},T^{\text{RL}}_{\text{pitch}},T^{\text{RL}}_{\text{yaw}})^{T}$
is a four dimensional vector that represents the residual of thrust and roll, pitch, yaw torques
and is computed as follows.
\begin{align}
	\label{eq4-1}
	a_{\pi_{\theta}}=
	\begin{cases}
		\sim \mathcal{N}(\mu_{\theta},\sigma_{\theta}) & \quad\text{(Training)} \\
		\mu_{\theta}                                   & \quad\text{(Testing)}
	\end{cases}
\end{align}
\begin{equation}
	\label{eq4-2}
	\hat{a_{\text{RL}}} = \tanh(a_{\pi_{\theta}})
\end{equation}
$\mu_{\theta}$ and $\sigma_{\theta}$ denotes the mean and the standard deviation of the Gaussian distribution computed with $\pi_{\theta}$.
$\hat{a_{\text{RL}}}$ is combined with PID-controller's output $a_{\text{PID}}$ as follows.
\begin{align}
	\label{eq4-3}
	 & a^{\text{shifted}}_{\text{RL}}=\frac{1}{2}(\hat{a_{\text{RL}}} + 1)\times\beta-\epsilon                                                             \\
	\label{eq4-4}
	 & a^{\text{scaled}}_{\text{RL}}=a^{\text{shifted}}_{\text{RL}}\odot(a^{\text{max}}_{\text{RL}}-a^{\text{min}}_{\text{RL}})+a^{\text{min}}_{\text{RL}} \\
	\label{eq4-5}
	 & a_{\text{RL}}=\text{clip}(a^{\text{scaled}}_{\text{RL}}, a^{\text{min}}_{\text{RL}}, a^{\text{max}}_{\text{RL}})                                    \\
	\label{eq4-6}
	 & a = a_{\text{PID}} + a_{\text{RL}}
\end{align}
Equation (\ref{eq4-3}) shifts the output squashed with $\tanh$ in (\ref{eq4-2}) to the range of $[-\epsilon, \beta-\epsilon]$.
This shift operation enables the reinforcement learning controller to output values between $[a^{\text{min}}_{\text{RL}}, a^{\text{max}}_{\text{RL}}]$
using (\ref{eq4-4}) and (\ref{eq4-5}). Without (\ref{eq4-3}), values near $a^{\text{min}}_{\text{RL}}$ and $a^{\text{max}}_{\text{RL}}$
are not computed because it is computationally infeasible for $\tanh$ to output values near -1 and 1.
$\odot$ denotes the hadamard product. In this work, we set $\beta=1.2$, $\epsilon=0.1$,
$a^{\text{max}}_{\text{RL}}=(0.25, 0.1, 0.1, 0.1)^{T}$
, and $a^{\text{min}}_{\text{RL}}=-a^{\text{max}}_{\text{RL}}$.
Finally, we compute propeller's angular velocity using the following relationship between
$a=(F_{\text{thrust}},T_{\text{roll}},T_{\text{pitch}},T_{\text{yaw}})^{T}$
in (\ref{eq4-6}) and propeller's angular velocity $\omega=(\omega_0, \omega_1, \omega_2, \omega_3)^{T}$.
\begin{align}
	\label{eq4-7}
	\begin{pmatrix}
		F_{\text{thrust}} \\
		T_{\text{roll}}   \\
		T_{\text{pitch}}  \\
		T_{\text{yaw}}
	\end{pmatrix}=
	\begin{pmatrix}
		1  & 1  & 1  & 1  \\
		-c & c  & c  & -c \\
		c  & -c & c  & -c \\
		1  & 1  & -1 & -1
	\end{pmatrix}
	\begin{pmatrix}
		{\omega_0}^2 \\
		{\omega_1}^2 \\
		{\omega_2}^2 \\
		{\omega_3}^2
	\end{pmatrix}, \quad c=\frac{1}{\sqrt{2}}
\end{align}

\subsection{Training Method}
\begin{figure}
	\centering
	\begin{minipage}[b]{0.49\columnwidth}
		\centering
		\includegraphics[width=\columnwidth]{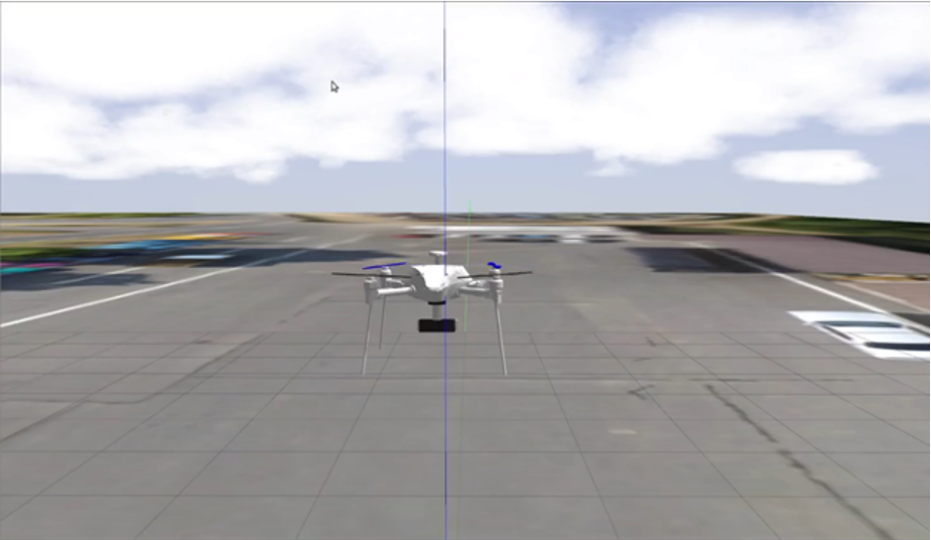}
	\end{minipage}
	\begin{minipage}[b]{0.425\columnwidth}
		\centering
		\includegraphics[width=\columnwidth]{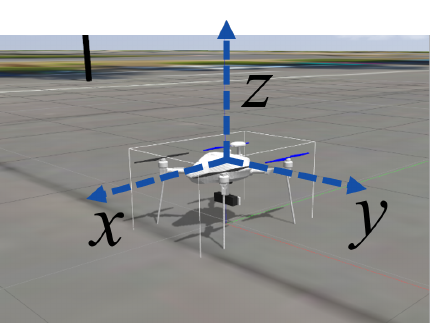}
	\end{minipage}
	\caption{Simulation environment.}
	\label{fig4-2}
\end{figure}
We trained the presented networks using SAC in a simulation environment shown in Fig. \ref{fig4-2}.
This simulation environment was built on top of Gazebo physics simulator~\cite{IROS2004_Gazebo}.
We used PX4~\cite{ICRA2015_Meier} as the base flight controller of our quadcopter~\cite{Aerosense2022_ASMC03T}
model in the simulation. We trained the controller using only a single quadcopter model.
However, we will show in the evaluation section that the trained controller also works
on quadcopters with mass and lift coefficient parameters different from original model.
During the training, simulated wind, between \SI{15}{m/s} to \SI{25}{m/s}, blows
from 26 different directions\footnote{$26=3^3-1$ comes from the possible combinations of $+,-,0$ for each axis. $(+, 0, 0), (+, +, 0), \dots, (+, +, +), (-, 0, 0), \dots, (-, -, -)$} at random.
The wind speed linearly increases from zero to selected wind speed in \SI{1}{s} and keeps blowing at selected speed for \SI{3}{s}.
We expected that, by training the quadcopter with different wind speed, the trained controller gets
robust not only against wind disturbances but also against changes in quadcopter's parameters (mass and lift coefficient).
This is because the controller trained to compensate for forces of different magnitudes should also
compensate for changes in the forces on the quadcopter due to differences in mass and lift-coefficient.

We designed the following reward function to train the network
\begin{equation}
	\label{eq4-8}
	r=w_{\text{pos}}r_{\text{pos}} + w_{\text{rp}}r_{\text{rp}} + w_{\text{yt}}r_{\text{yt}}.
\end{equation}
$w_{\text{pos}}$, $w_{\text{rp}}$, and $w_{\text{yt}}$ are the weights of each term and
$r_{\text{pos}}$, $r_{\text{rp}}$, and $r_{\text{yt}}$ are described as follows:
\begin{align}
	\label{eq4-9}
	 & r_{\text{pos}} = \min(-\Delta_{t}+\Delta_{t-1}, 0),                                                            \\
	 & \text{where}\ \Delta_t=\sqrt{\Delta x_{t}^{2}+\Delta y_{t}^{2}+\Delta z_{t}^{2}} \nonumber                     \\
	\label{eq4-10}
	 & r_{\text{rp}} = -(|T_{\text{roll}, t} -T_{\text{roll}, t-1}| + |T_{\text{pitch}, t} - T_{\text{pitch}, t-1}|)  \\
	\label{eq4-11}
	 & r_{\text{yt}} = -(|T_{\text{yaw}, t} - T_{\text{yaw}, t-1}| + |F_{\text{thrust}, t} -F_{\text{thrust}, t-1}|).
\end{align}
$\Delta x_{t}$, $\Delta y_{t}$, and $\Delta z_{t}$ denotes the relative distance to the target position for each axis at time $t$.
Hence, $r_{\text{pos}}$ penalizes the policy when the quadcopter diverges from the target position compared to previous timestep.
$r_{\text{rp}}$ and $r_{\text{yt}}$ are designed for similar purpose to penalize rapid changes in the output of policy $\pi_\theta$.
We weighted (\ref{eq4-10}) and (\ref{eq4-11}) separately because we assumed that roll and pitch angle affect
quadcopter's position greater than yaw angle and thrust. We set the weights to $w_{\text{pos}}=1$, $w_{\text{rp}}=2.5\times10^{-2}$,
and $w_{\text{yt}}=5\times10^{-3}$.

We set the discount factor $\gamma$ to 0.95, batch size to 128, and target network update coefficient $\tau$ used in SAC to $1.0\times10^{-4}$.
Network parameters were updated using Adam~\cite{ICLR2015_Kingma} with learning rate of $2.0\times10^{-4}$.
During the training, we applied uniformly sampled noise between $[-1, 1)$ to each state element in batch
to imitate the sensor noise on real quadcopters.
When applying the noise, we scaled the sampled noise
by 0.1 for relative position,
by 0.5 for velocity,
by 0.05 for relative angle,
by 1.25 for angular velocity, and
by 0.1 for PID-controller's output.
We did not append this noise to RL-controller's output.
We ran the training until average reward received by the quadcopter converges and used the best parameter learned in the evaluation.
The training took approximately \SI{12}{hours} (wall-clock time).

\section{SIMULATION}

\subsection{Performance Evaluation}
We evaluated the performance of the proposed controller in the simulation environment presented in previous section (See Fig.~\ref{fig4-2}).
During the evaluation, simulated gust of wind of \SI{20}{m/s} blows from 26 different directions as in previous section
to check the performance of trained controller.
For each wind direction, the wind blows once for \SI{30}{steps} (i.e. \SI{3}{s}).
There are \SI{70}{steps} (i.e. \SI{7}{s}) of interval until the wind blows from another direction.
In the evaluation, we used the conventional cascaded PID controller as a baseline method and compared with the proposed controller.
In prior to the evaluation, the PID parameters were manually tuned using real quadcopter (AS-MC03-W2~\cite{Aerosense2022_ASMC03W},
see Appendix \ref{app:hardware-spec}), and selected the best performing PID parameters under wind disturbances for the comparison.
We used this PID parameters for both evaluation and training of the proposed controller.

\begin{figure}
    \centering
    \begin{minipage}{1.0\columnwidth}
        \centering
        \includegraphics[width=\columnwidth]{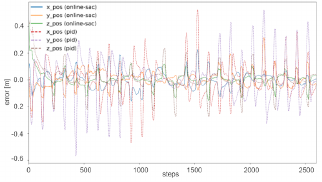}
    \end{minipage}
    \caption{Positional error of quadcopter during evaluation (Dotted lines: conventional cascaded PID-controller, solid lines: PID+reinforcement learning controller (Proposed)).
        Proposed controller reduces the deviation by approximately \SI{50}{\%} compared to PID-only controller.}
    \label{fig5-1}
\end{figure}

Fig.~\ref{fig5-1} shows the position history of the quadcopter during the simulation.
Each of the 26 different positional peak in the figure corresponds to the number of wind directions blown in the evaluation.
Proposed controller succeeds in reducing the deviation in all wind directions by approximately \SI{50}{\%} compared to the baseline cascaded PID controller.
The overall result is shown in Table~\ref{tab5-1}.
From Table~\ref{tab5-1}, we can also verify that the proposed controller succeeds in reducing the deviation from target position.
The improvement in standard deviation of overall positional error was around \SI{45}{\%} to \SI{63}{\%} in our evaluation.

\begin{table}[t]
    \caption{Positional error of each controller.}
    \label{tab5-1}
    \centering
    \begin{tabular}{|c|c|c|c|}
        \hline
        \multirow{2}*{Axis} & Cascaded PID [m]    & Proposed [m]        & \multirow{2}*{Improvement} \\
                            & (mean$\pm$std dev.) & (mean$\pm$std dev.) &                            \\
        \hline
        x                   & $0.003\pm0.132$     & $0.002\pm0.060$     & \SI{55}{\%}                \\
        \hline
        y                   & $0.002\pm0.166$     & $0.002\pm0.063$     & \SI{63}{\%}                \\
        \hline
        z                   & $0.014\pm0.099$     & $0.006\pm0.055$     & \SI{45}{\%}                \\
        \hline
    \end{tabular}
\end{table}

\subsection{Robustness Evaluation}
We checked the robustness of proposed controller against the changes in quadcopter's mass and propeller's lift coefficient.
We evaluated the robustness against these changes because quadcopter's mass may change dynamically when it loads or unloads additional payloads and
propeller's lift coefficient is affected by the air density of the environment.
In addition, these parameters change depending on the hardware.
We used the same controller and same experiment procedure used in the performance evaluation for this robustness evaluation.
During the evaluation, we fixed the lift coefficient to original value and
changed the mass of the quadcopter and vice versa to check the robustness of the controller against each parameter.

\begin{figure}[t]
    \centering
    \begin{minipage}{0.8\columnwidth}
        \centering
        \includegraphics[width=\columnwidth]{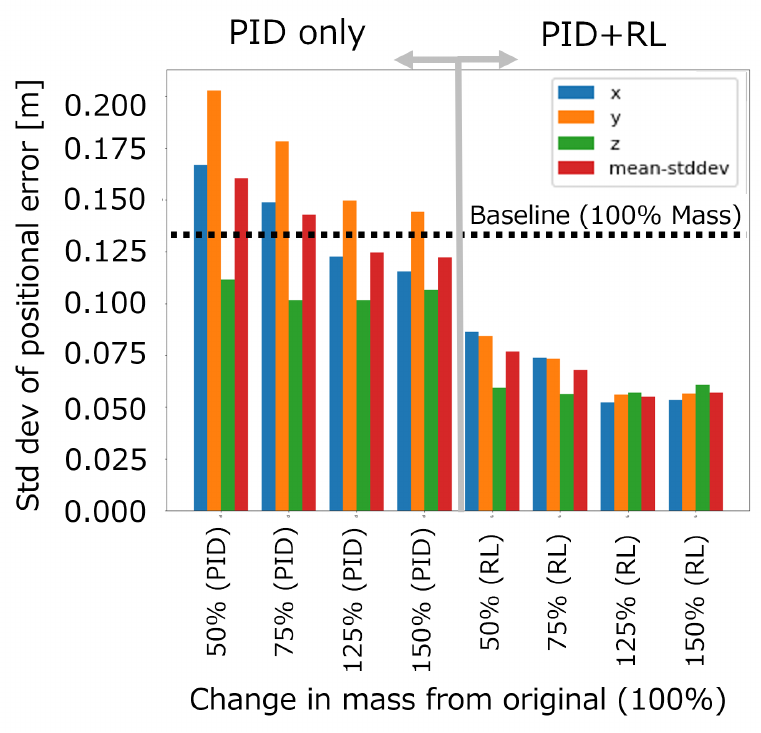}
        \subcaption{Mass of the quadcopter vs positional error}
        \label{fig5-2}
    \end{minipage}\\
    \begin{minipage}{0.8\columnwidth}
        \centering
        \includegraphics[width=\columnwidth]{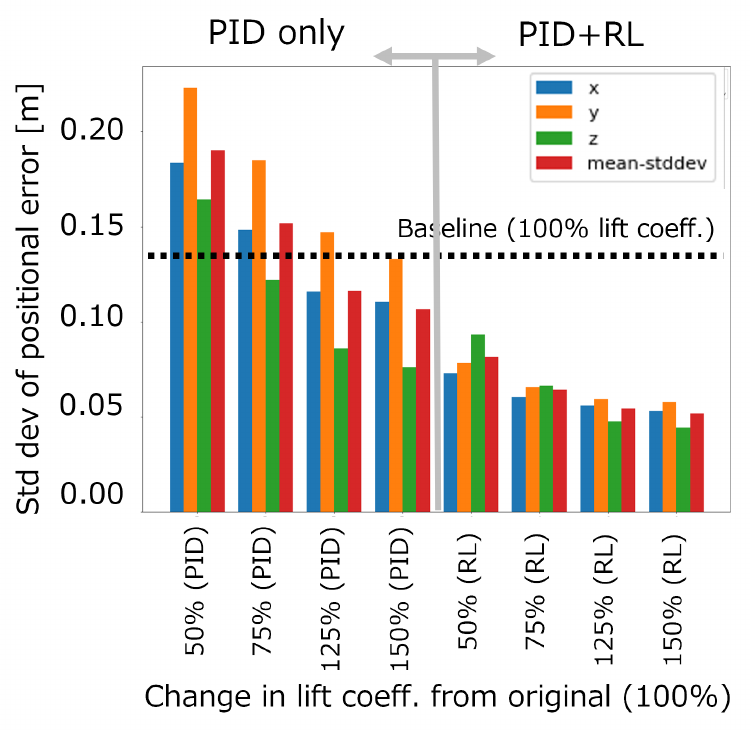}
        \subcaption{Propeller's lift coefficient vs positional error}
        \label{fig5-3}
    \end{minipage}
    \caption{Relationship between the positional error of quadcopter and changes in quadcopter's parameters.
        Dotted line shows the baseline PID controller's performence.
        Reinforcement learning controller preserves its performance even though the parameters change.}
\end{figure}
Fig.~\ref{fig5-2} and Fig.~\ref{fig5-3} show the standard deviation of the quadcopter's position against the changes in its parameters.
Dottled line shows the average standard deviation of the quadcopter with original mass and lift coefficient controlled using conventional cascaded PID-controller.
From the figure, we can confirm that the proposed controller preserves its performance even though the
quadcopter's parameter changes from original training time.
The proposed controller succeeded in reducing the deviation from target position by approximately \SI{40}{\%} to \SI{60}{\%}
compared to the conventional cascaded PID controller. Even though the quadcopter's mass or propeller's lift coefficient changes drastically
(in the range of \SI{50}{\%} to \SI{150}{\%}), the proposed controller performs better than manually tuned PID-controller on
original quadcopter.
See also Appendix~\ref{app:extra-simulation-result} for the performance of
proposed controller on other combinations of mass and lift coefficient values.

\section{OUTDOOR EXPERIMENTS}
\begin{figure}
	\centering
	\begin{minipage}{0.485\columnwidth}
		\centering
		\includegraphics[width=\columnwidth]{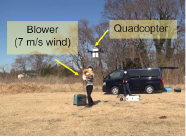}
		\subcaption{Low wind speed}
		\label{fig6-1a}
	\end{minipage}
	\begin{minipage}{0.49\columnwidth}
		\centering
		\includegraphics[width=\columnwidth]{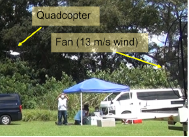}
		\subcaption{High wind speed}
		\label{fig6-1b}
	\end{minipage}
	\caption{Outdoor experimental environment}
	\label{fig6-1}
\end{figure}
We evaluated the performance of proposed controller using a real quadcopter in an outdoor environment shown in Fig.~\ref{fig6-1}.
The evaluation was performed in two different conditions: low wind speed (less than \SI{7}{m/s}) condition and high wind speed (greater than \SI{13}{m/s}) condition.
We evaluated the same controller used in the simulation without extra fine tuning for this experiment.

\subsection{Performance under Low Wind Speed Condition}
\begin{figure*}[tb]
	\centering
	\begin{minipage}{1.0\columnwidth}
		\centering
		\begin{minipage}{0.49\columnwidth}
			\centering
			\includegraphics[width=\columnwidth]{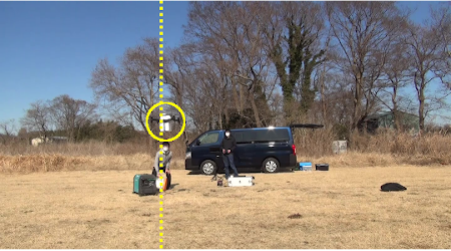}
			\subcaption{\SI{0.0}{s}}
			\label{fig6-2a}
		\end{minipage}
		\begin{minipage}{0.49\columnwidth}
			\includegraphics[width=\columnwidth]{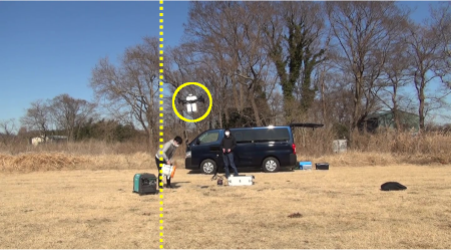}
			\subcaption{\SI{12.5}{s}}
			\label{fig6-2b}
		\end{minipage} \\
		\begin{minipage}{0.49\columnwidth}
			\centering
			\includegraphics[width=\columnwidth]{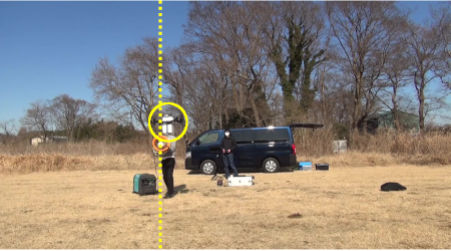}
			\subcaption{\SI{5.0}{s}}
			\label{fig6-2c}
		\end{minipage}
		\begin{minipage}{0.49\columnwidth}
			\includegraphics[width=\columnwidth]{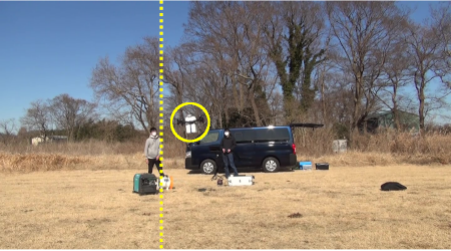}
			\subcaption{\SI{15.0}{s}}
			\label{fig6-2d}
		\end{minipage} \\
		\begin{minipage}{0.49\columnwidth}
			\centering
			\includegraphics[width=\columnwidth]{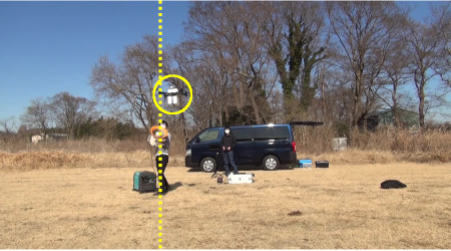}
			\subcaption{\SI{10.0}{s}}
			\label{fig6-2e}
		\end{minipage}
		\begin{minipage}{0.49\columnwidth}
			\includegraphics[width=\columnwidth]{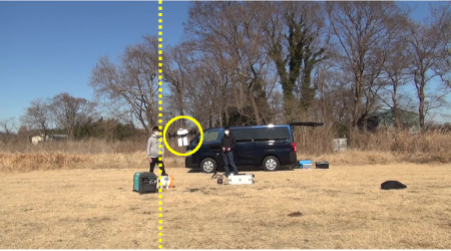}
			\subcaption{\SI{20.0}{s}}
			\label{fig6-2f}
		\end{minipage}
		\caption{Snapshots of the quadcopter with cascaded PID-controller under low wind speed Condition.}
		\label{fig6-2}
	\end{minipage}
	\begin{minipage}{1.0\columnwidth}
		\centering
		\begin{minipage}{0.49\columnwidth}
			\centering
			\includegraphics[width=\columnwidth]{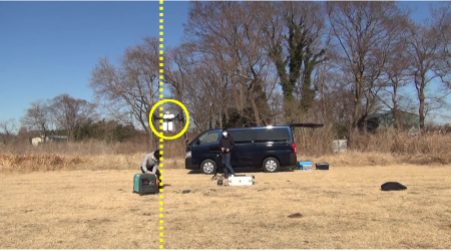}
			\subcaption{\SI{0.0}{s}}
			\label{fig6-3a}
		\end{minipage}
		\begin{minipage}{0.49\columnwidth}
			\includegraphics[width=\columnwidth]{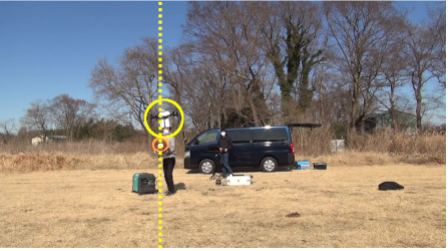}
			\subcaption{\SI{12.5}{s}}
			\label{fig6-3b}
		\end{minipage} \\
		\begin{minipage}{0.49\columnwidth}
			\centering
			\includegraphics[width=\columnwidth]{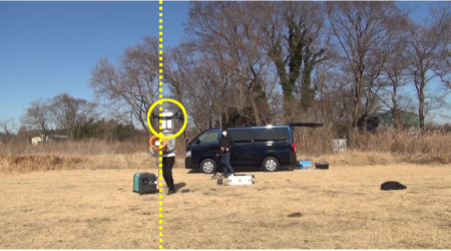}
			\subcaption{\SI{5.0}{s}}
			\label{fig6-3c}
		\end{minipage}
		\begin{minipage}{0.49\columnwidth}
			\includegraphics[width=\columnwidth]{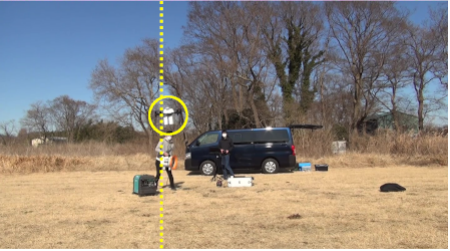}
			\subcaption{\SI{15.0}{s}}
			\label{fig6-3d}
		\end{minipage} \\
		\begin{minipage}{0.49\columnwidth}
			\centering
			\includegraphics[width=\columnwidth]{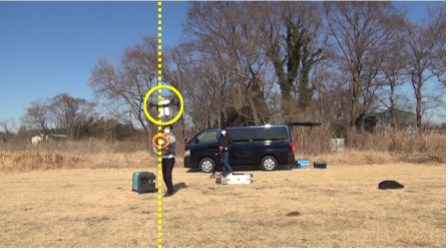}
			\subcaption{\SI{10.0}{s}}
			\label{fig6-3e}
		\end{minipage}
		\begin{minipage}{0.49\columnwidth}
			\includegraphics[width=\columnwidth]{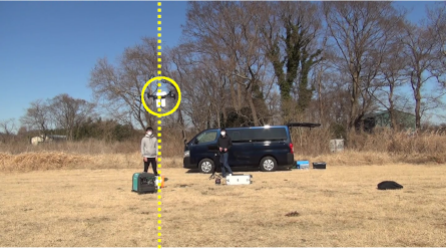}
			\subcaption{\SI{20.0}{s}}
			\label{fig6-3f}
		\end{minipage}
		\caption{Snapshots of the quadcopter with proposed controller under low wind speed condition.}
		\label{fig6-3}
	\end{minipage}
\end{figure*}

\begin{table}[t]
	\caption{Average and maximum absolute positional error of quadcopter under low wind speed condition.}
	\label{tab6-1}
	\centering
	\begin{tabular}{|c|c|c|c|}
		\hline
		                    & Cascaded PID                 & Proposed                     & Improvement           \\
		\hline
		Average [m]         & \multirow{2}*{$0.29\pm0.17$} & \multirow{2}*{$0.17\pm0.09$} & mean: \SI{41}{\%}     \\
		(mean$\pm$std dev.) &                              &                              & std dev.: \SI{53}{\%} \\
		\hline
		Maximum [m]         & 0.93                         & 0.48                         & \SI{52}{\%}           \\
		\hline
	\end{tabular}
\end{table}
Fig.~\ref{fig6-2} and Fig.~\ref{fig6-3} show the behavior of the quadcopter during the experiment.
For this experiment, we used a quadcopter with an extra payload of \SI{1}{kg}.
See Appendix~\ref{app:hardware-spec} for the details of the hardware used.
In the experiment, we manually hit the wind using a blower for around \SI{5}{s}-\SI{10}{s} several times and checked the behavior of the quadcopter.
From Fig.~\ref{fig6-2}, we can confirm that the conventional cascaded PID controller easily deviates from the target position
when the wind from the blower hits the quadcopter. Conversely, from Fig.~\ref{fig6-3}, we can verify that
the proposed controller succeeds in maintaining its position.
\begin{figure}[!tb]
	\centering
	\includegraphics[width=\columnwidth]{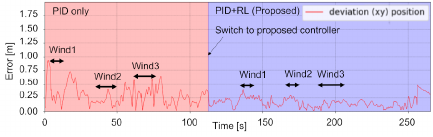}
	\caption{History of absolute positional error of the quadcopter during low wind speed experiment.
		Left and right arrows show the wind blowing times.}
	\label{figA-3}
\end{figure}
Fig.~\ref{figA-3} shows the full history of absolute positional error of the quadcopter during this experiment.
We hit the wind to the quadcopter three times each.
From the figure, we can also verify that the proposed controller is effective in stabilizing the quadcopter around target position.
Table~\ref{tab6-1} shows the overall result of the experiment.
The mean and standard deviation in the table shows the average deviation in the xy-coordinate.
We removed the z-coordinate from the calculation because we used a barometer-based sensor to compute the quadcopter's position in the z-coordinate.
Barometer-based localization was inaccurate when the gust of wind hit the quadcopter.
From Table~\ref{tab6-1}, we can confirm that the deviation from target position using the proposed controller is smaller than the
conventional cascaded PID controller.
Our controller succeeded in reducing the deviation by \SI{41}{\%} on average and by \SI{52}{\%} on maximum position
compared to the conventional cascaded PID controller.

\subsection{Performance under High Wind Speed Condition}
In this experiment, we used a large fan shown in Fig.~\ref{fig6-1b} to imitate a scene with complex wind gust condition.
We used a different quadcopter from previous evaluation to check the robustness of our controller.
The quadcopter has different shapes and is connected by a wire for electric power supply
(i.e. extra load is applied depending on the altitude of the quadcopter).
See also Appendix~\ref{app:hardware-spec} for the details of the hardware used.
The fan shown in Fig.~\ref{fig6-1b} continuously sends wind of approximately \SI{13}{m/s} to the quadcopter during the experiment.
We limited the wind speed to \SI{13}{m/s} considering the safety of the quadcopter during the experiment.
According to the Beaufort scale~\cite{NWS2023_Beaufort},
wind speed greater than \SI{13}{m/s} is classified as an environment where
people feel inconvenience walking against the wind.
The wind gust is highly difficult to predict because the fan generates a complex current of air around the quadcopter.
During the experiment, we monitored the source wind speed and kept the source wind speed greater than \SI{13}{m/s}.

\begin{figure}
	\centering
	\begin{minipage}{0.95\columnwidth}
		\centering
		\includegraphics[width=\columnwidth]{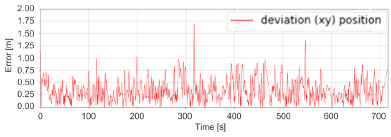}
		\subcaption{Cascaded PID controller}
		\label{fig6-4a}
	\end{minipage} \\
	\begin{minipage}{0.95\columnwidth}
		\centering
		\includegraphics[width=\columnwidth]{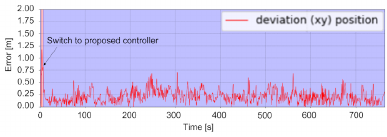}
		\subcaption{Proposed controller}
		\label{fig6-4b}
	\end{minipage}
	\caption{History of absolute positional error of the quadcopter during high wind speed experiment.
		The wind was blowing constantly throughout the experiment.}
	\label{fig6-4}
\end{figure}

Fig.~\ref{fig6-4} shows the history of quadcopter's positional error during the experiment.
From the figure, we can confirm that the proposed controller significantly reduces the deviation from target position
compared to the conventional cascaded PID controller.
Convetional controller deviated from target position by more tham \SI{1}{m} several times during the experiment.
In contrast, proposed controller successfully stabilized the quadcopter and prevented the quadcopter largely deviating from target position.
The maximum deviation of proposed controller was approximately \SI{0.78}{m}.
Table \ref{tab6-2} shows the overall result of the experiment.
From the table, we can also verify that the proposed controller improves the stabilization performance of the quadcopter.
Proposed controller reduces the deviation from target position by \SI{35}{\%} on average and by \SI{53}{\%} on maximum position.

\begin{table}[t]
	\caption{Average and maximum absolute positional error of quadcopter under high wind speed condition.}
	\label{tab6-2}
	\centering
	\begin{tabular}{|c|c|c|c|}
		\hline
		                    & Cascaded PID                 & Proposed                     & Improvement           \\
		\hline
		Average [m]         & \multirow{2}*{$0.34\pm0.20$} & \multirow{2}*{$0.22\pm0.13$} & mean: \SI{35}{\%}     \\
		(mean$\pm$std dev.) &                              &                              & std dev.: \SI{33}{\%} \\
		\hline
		Maximum [m]         & 1.69                         & 0.78                         & \SI{53}{\%}           \\
		\hline
	\end{tabular}
\end{table}

\section{CONCLUSION AND FUTURE WORK}

We presented a residual reinforcement learning approach to build a wind resistance controller for quadcopters.
Proposed method uses conventional cascaded PID-controller as a base controller and learns the
residual input that compensates the wind disturbances.
The residual controller can be trained using only a simulator and no extra finetuning is necessary to run on a real hardware.
We evaluated the proposed controller's performance in both simulation and experiment in a real environment.
The trained controller reduces the positional deviation by approximately \SI{50}{\%} compared to
conventional cascaded PID controller. In addition, the trained controller preserves its performance
even though the quadcopter's parameters (mass and propeller's lift coefficient) have changed from original training time
in the range of \SI{50}{\%} to \SI{150}{\%}.

We demonstrated that our controller works not only in simulation but also on real hardware.
During the experiment, we could not observe the system getting unstable.
However, the stability of the proposed system is not guaranteed theoretically and may get unstable in a special situation.
Investigating the stability of the system could be a future work of this study.

\appendices
\section{Hardware Specification}
\label{app:hardware-spec}
\begin{figure}[!tb]
    \centering
    \begin{minipage}{0.40\columnwidth}
        \centering
        \includegraphics[width=\columnwidth]{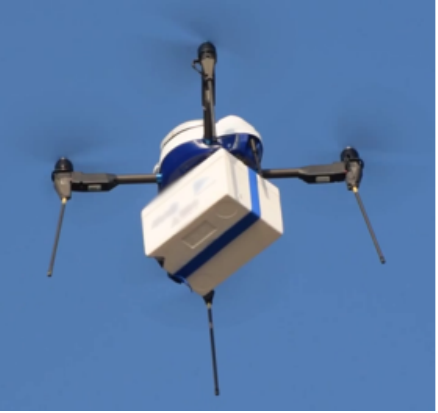}
        \subcaption{AS-MC03-T}
        \label{figA-1a}
    \end{minipage}
    \begin{minipage}{0.485\columnwidth}
        \centering
        \includegraphics[width=\columnwidth]{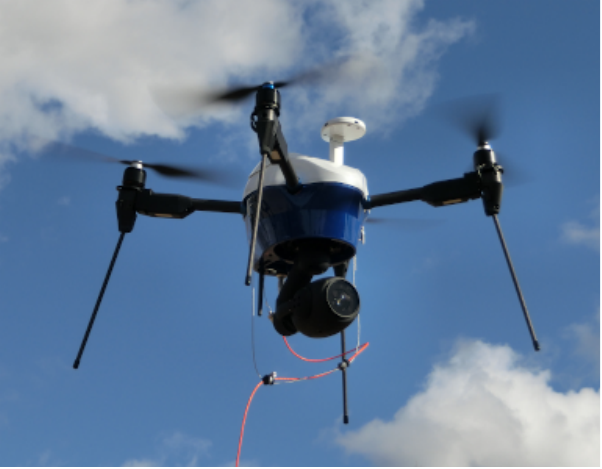}
        \subcaption{AS-MC03-W2}
        \label{figA-1b}
    \end{minipage}
    \caption{Quadcopters used in each experiment.
        (a) used in low wind speed experiment.
        (b) used in high wind speed experiment.}
    \label{figA-1}
\end{figure}

Fig.~\ref{figA-1} shows the quadcopter used in each real environment experiment~\cite{Aerosense2022_ASMC03W},~\cite{Aerosense2022_ASMC03T}.
See Table~\ref{tabA-2} and Table~\ref{tabA-3} for the specification of each quadcopter.
Please note that AS-MC03-T is a wireless quadcopter and AS-MC03-W2 is a wired quadcopter.
Therefore a extra load is applied to AS-MC03-W2 during the flight depending on the flight altitude.

\begin{table}[t][!tb]
    \caption{Hardware Specification of AS-MC03-T.}
    \label{tabA-2}
    \centering
    \begin{tabular}{|c|c|}
        \hline
        Size (length$\times$width$\times$height) & 943 mm$\times$943 mm$\times$450 mm \\
        \hline
        Weight                                   & \SI{3.53}{kg}                      \\
        \hline
        Max payload                              & \SI{3}{kg}                         \\
        \hline
        Max velocity                             & \SI{54}{km/h}                      \\
        \hline
        Shipping box size                        & 186 mm$\times$258 mm$\times$155 mm \\
        \hline
        Shipping box weight                      & \SI{267}{g}                        \\
        \hline
    \end{tabular}
\end{table}

\begin{table}[!tb]
    \caption{Hardware Specification of AS-MC03-W2.}
    \label{tabA-3}
    \centering
    \begin{tabular}{|c|c|}
        \hline
        Size (length$\times$width$\times$height) & 943 mm$\times$943 mm$\times$450 mm            \\
        \hline
        Weight (Quadcopter + Camera)             & \SI{5.12}{kg} (\SI{4.14}{kg} + \SI{0.98}{kg}) \\
        \hline
        Maximum payload                          & \SI{3}{kg}                                    \\
        \hline
        Maximum velocity                             & \SI{18}{km/h}                                 \\
        \hline
    \end{tabular}
\end{table}

\section{Extra Simulation Results}
\label{app:extra-simulation-result}
\begin{figure}[!tb]
    \centering
    \begin{minipage}{0.98\columnwidth}
        \includegraphics[width=\columnwidth]{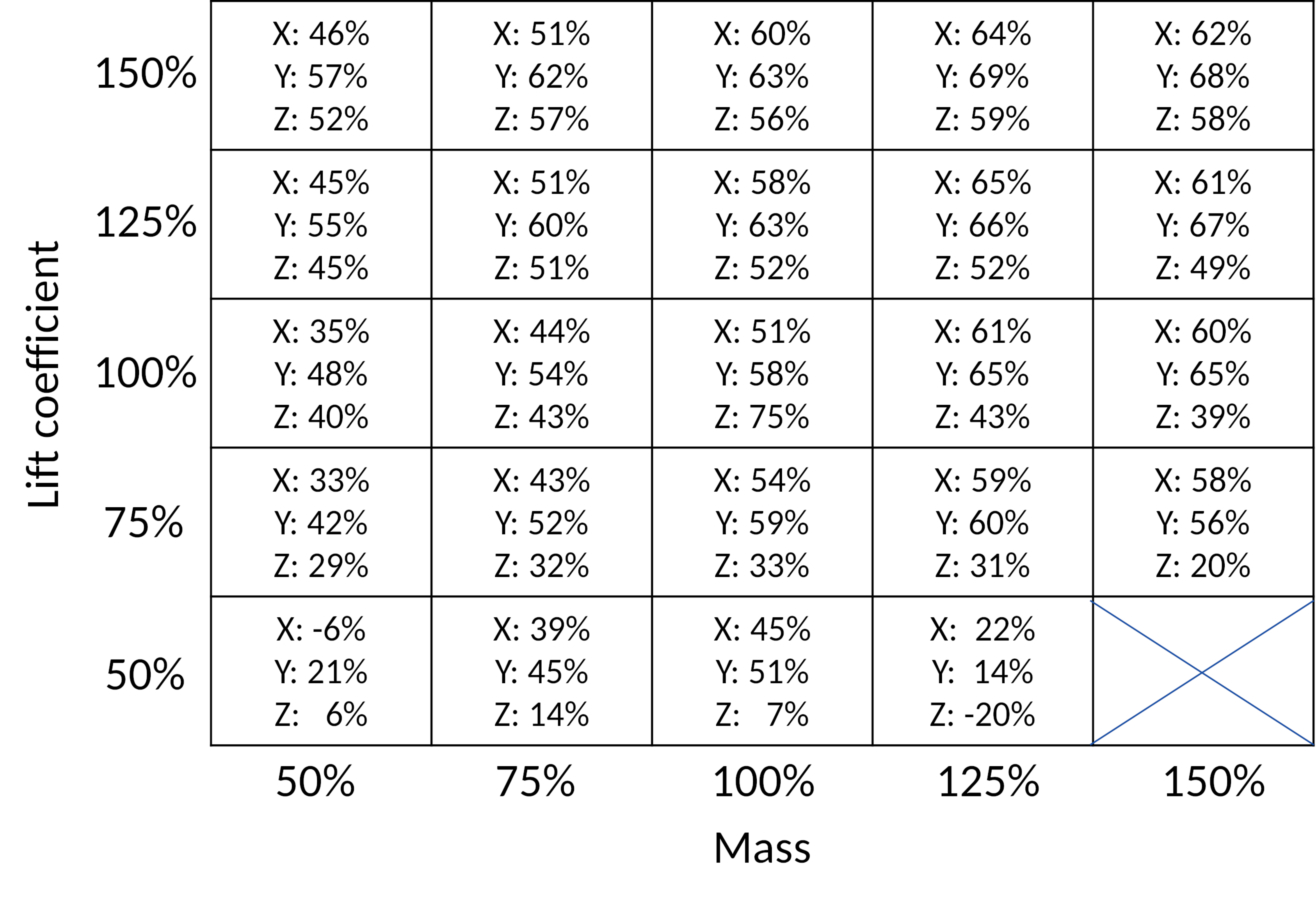}
    \end{minipage}
    \caption{Improvement in wind resistance performance along each axis.
        For x and y axis, the more the mass and lift coefficient, the more improvement in performance.
        For z axis, original weight and lift coefficient performs best.
        Box with a cross denotes the parameter that failed to fly.}
    \label{figA-2}
\end{figure}

Matrix on Fig.~\ref{figA-2} shows the performance improvement of
the proposed controller against the conventional cascaded PID-controller run on original quadcopter (\SI{100}{\%} mass and lift coefficient).
We can confirm that the propsed controller improves the control performance in most of the combinations by
\SI{50}{\%} or greater.

\section*{Acknowledgment}
We would like to express our special thanks to Hirotaka Suzuki, Shunichi Sekiguchi, and Tokuhiro Nishikawa at Sony Group Corporation
for their helpful feedback on the early draft of this manuscript and Aerosense Inc. members for their kind support on conducting the experiments.

\addtolength{\textheight}{-14cm}
\bibliographystyle{IEEEtran}
\bibliography{ref_iros2023}

\end{document}